\def\year{2024}\relax
\documentclass[letterpaper]{article} 
\usepackage{aaai22}  
\usepackage{times}  
\usepackage{helvet}  
\usepackage{courier}  
\usepackage[hyphens]{url}  
\usepackage{graphicx} 
\usepackage{natbib}  
\usepackage{caption} 
\DeclareCaptionStyle{ruled}{labelfont=normalfont,labelsep=colon,strut=off} 
\usepackage{newfloat}
\usepackage{listings}
\usepackage{algorithm}
\usepackage{algorithmicx}
\usepackage{algpseudocode}
\usepackage{amsmath,amsfonts}
\usepackage{amsmath}
\usepackage[utf8]{inputenc} 
\usepackage[T1]{fontenc}    
\usepackage{hyperref}       
\usepackage{url}            
\usepackage{booktabs}       
\usepackage{amsfonts}       
\usepackage{amsthm}
\usepackage{nicefrac}       
\usepackage{microtype}      
\usepackage{xcolor}         
\usepackage{color,soul}

\usepackage{bbm}
\usepackage{wrapfig}
\usepackage{caption}
\usepackage{subcaption}

\definecolor{forestgreen}{rgb}{0.13, 0.55, 0.13}
\newtheorem{thm}{Theorem}[section] 

\newtheorem{defn}[thm]{Definition} 

\floatstyle{ruled}
\newfloat{listing}{tb}{lst}{}
\floatname{listing}{Listing}

\title{Evaluating Privacy Leakage in Split Learning}
\author {
    Xinchi Qiu,\textsuperscript{\rm 1} \footnote{This work is done while Xinchi Qiu was interning at Meta, Email: \url{xq227@cam.ac.uk}}
    Ilias Leontiadis \textsuperscript{\rm 2}
    Luca Melis \textsuperscript{\rm 2}
    Alex Sablayrolles \textsuperscript{\rm 2}
    Pierre Stock \textsuperscript{\rm 2}
}
\affiliations {
    \textsuperscript{\rm 1} University of Cambridge\\
    \textsuperscript{\rm 2} Meta\\
}

\usepackage{bibentry}

\begin{document}

\maketitle

\begin{abstract}
Privacy-Preserving machine learning (PPML) can help us train and deploy models that utilize private information. In particular, on-device machine learning allows us to avoid sharing raw data with a third-party server during inference. On-device models are typically less accurate when compared to their server counterparts due to the fact that (1) they typically only rely on a small set of on-device features and (2) they need to be small enough to run efficiently on end-user devices.
Split Learning (SL) is a promising approach that can overcome these limitations. In SL, a large machine learning model is divided into two parts, with the bigger part residing on the server side and a smaller part executing on-device, aiming to incorporate the private features. However, end-to-end training of such models requires exchanging gradients at the cut layer, which might encode private features or labels. In this paper, we provide insights into potential privacy risks associated with SL. Furthermore, we also investigate the effectiveness of various mitigation strategies. Our results indicate that the gradients significantly improve the attacker’s effectiveness in all tested datasets reaching almost perfect reconstruction accuracy for some features. However, a small amount of differential privacy (DP) can effectively mitigate this risk without causing significant training degradation. 
\end{abstract}

\section{Introduction}

On-device machine learning (ML) involves training and/or deploying models directly on the device without relying on cloud-based computing. This approach brings several benefits to the table, including increased privacy, reduced latency, and access to fine-grained real-time data. Such models have been deployed for a variety of machine learning tasks, such as smart keyboard \cite{abdulrahman2020survey}, personalized assistant services \cite{hao2020apple}, computer vision \cite{liu2020fedvision}, healthcare \cite{rieke2020future}, and ranking \cite{fel,hartmann2019federated, paulik2021federated}.

At the same time, on the other hand, there are certain limitations that hinder the wide adoption of on-device AI models. Firstly, the limited computational and memory resources of client devices also restrict the size of the deployed models. As a result, the learning capacity and accuracy can be significantly worse than the equivalent server-based models. Secondly, end-user devices might have limited access to large datasets or have the capacity to store and process features that require large embedding tables. 

While on-device AI helps us ensure privacy, a key observation is that not all features might be sensitive, user-specific, or generated on-device. Examples include e-commerce item embeddings in a recommendation system, word embeddings of a large language model, ads-related features, etc. As such, training a small model entirely on-device might not be the optimal policy.

One promising approach to overcome these limitations is Split Learning (SL)~\cite{gupta2018distributed, vepakomma2020splintering}. Typically, a large machine learning model is divided into two parts: the bigger part resides on the server side (typically hosted by the model owner), and a smaller part can be executed on-device (typically hosted by the end-users). Larger models can then be collaboratively trained on both private (client) and non-private (server) features while limiting the information exchange between the involved parties. 

During inference, the server initiates the forward pass utilizing all the server-side features. Large embedding tables and model architectures can be utilized at this stage. Only the activations of the \emph{cut layer} are then shared with the end devices, typically a small vector. Each device continues the execution on its own sub-model, combining it with its own private features. Due to the limited capabilities, the client-side model is small, often only utilizing numerical features or categorical features with small cardinality. In this paper, we consider the most private scenario where the label is also private (e.g., the label represents a user purchase or a conversion after seeing an ad).

While split learning only considers two parties (e.g., ads publishers and advertisers),  Federated Split Learning (FSL) allows us to train such models between a central party and millions of client devices \cite{thapa2022splitfed}. 
In both cases, training the server-side model requires exchanging the gradients at the cut layer for each sample. Consequently, the returning cut-layer gradients might encode information that can reveal either private features and/or labels.

This paper provides insights into the potential risks of private data leakage during split model training. To achieve this, we introduce an extensive attack method that aims to reconstruct private information - features or labels - by exploiting diverse information sources at the cut layer, such as the model parameters, the activations, and gradients at the cut layer. Through our study, we aim to highlight the potential privacy risks associated with split learning by studying in which way features might be more sensitive to leakage. Finally, a significant part of the paper is devoted to studying how different strategies can help us mitigate these risks.

To sum up, in this work:
\begin{itemize}

    \item We are the first to use an extensive and custom clustering-based attack method on SL. This method exploits diverse information sources, such as the model parameters, activations, and gradients at the cut layer, to reconstruct private information, including features or labels.
    \item We study how different mitigation strategies, such as label and gradient differential privacy (DP), can help us protect such on-device private features. 
    \item Our results over three different datasets indicate that the gradients significantly improve the effectiveness of the attack methods when compared to the baselines. In all three datasets, an attacker is able to perfectly reconstruct labels and most features. However, adding a small amount of noise on the gradients at the cut layer (e.g., $\sigma=0.01$) effectively mitigates this risk with a mere $0.01$ drop in the model's AUC.
\end{itemize}
\section{Background and Related Work}\label{sec:related}

\vspace{3mm}

\paragraph{Split Learning (SL)} enables collaborative training of deep learning models among multiple parties without the need to share raw data. While Federated Learning~\cite{fedavg} can be utilized for such models, it may not always be practical. For instance, e-commerce and ad-ranking models often involve numerous sparse (categorical) and numerical features, requiring large machine learning models that can reach sizes of hundreds of gigabytes. Such models are typically too large to be trained on mobile devices \cite{qiu2022zerofl, horvath2021fjord}, whereas user-side features and labels (e.g., past purchases) might be too sensitive to collect on the server side. 

\begin{figure*}[t]
    \centering
    \includegraphics[width=0.9\textwidth]{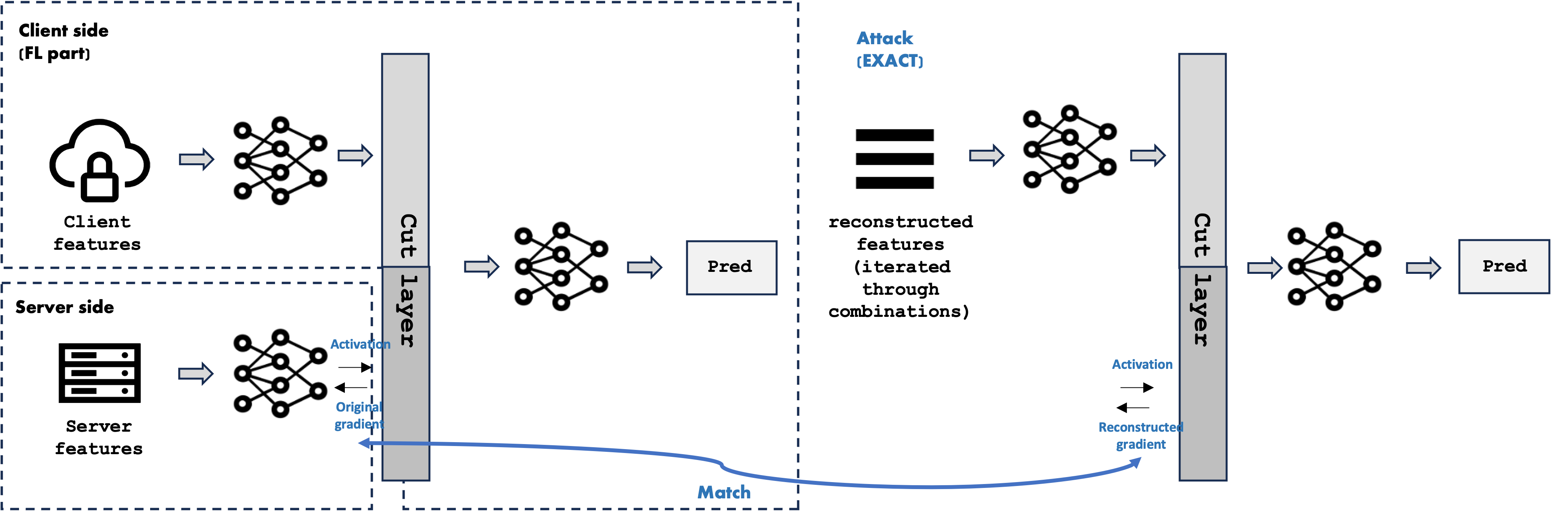}
    \caption{Illustration of SL (left) and our attack (right). During training, the server performs a forward pass until the cut later and then sends the intermediate representations to each client. This information and private features are used on-device to resume the computation. During the backward pass, the partial gradients returned might encode private client features or labels. Our attack uses these gradients to reconstruct the private features. }
   \label{fig:fig}
\end{figure*}

In split learning (SL), the overall model is horizontally divided into two parts. The server handles the forward/backward passes of the first and larger portion of the model while keeping the last few layers to be trained with sensitive user-side data/features on the device. This division point in the model architecture is referred to as the \emph{cut layer}. The server uses server-side features to perform a forward pass until the cut layer and then forward the intermediate representations to the respective clients. At each client, a smaller architecture processes the private features, which are then combined with the server-side representations. An overarching architecture is employed to make the final prediction. An example illustrating this process is depicted on the left side of Figure~\ref{fig:fig}. During back-propagation, gradients are calculated from the last layer to the cut layer in a similar manner. The corresponding gradients are then sent back to the server to complete the server-side back-propagation. While no raw data are exchanged, these gradients might still encode private information that can reveal either features and/or the ground truth labels that are stored on the devices. 

\paragraph{Membership inference attacks} aim at inferring whether a given sample was part of a model's training set. Originally proposed in~\cite{homer2008resolving}, it was popularized by the shadow models approach of Shokri et al.~\cite{shokri2017membership}.
Follow-up work has shown that the membership signal can be derived from simple quantities, such as a low loss~\cite{salem2018ml,sablayrolles2019white}. Recent research has improved the performance of such attacks by comparing the loss to a calibrating term~\cite{watson2021importance} or computing statistics on the distribution of the loss~\cite{carlini2022membership}.

\paragraph{Reconstruction attacks and attribute inference} aim at reconstructing points from the training set given access to a trained model~\cite{carlini2019secret}, and to partial information about the sample in the case of attribute inference~\cite{fredrikson2014privacy,yeom2018privacy}.
Carlini et al.~\cite{carlini2021extracting} show that given access to a trained language model, an attacker is able to reconstruct verbatim samples with high precision (but low recall).

\paragraph{Attacks on Federated Learning.} Recovering private features from gradients has gained growing interest in the privacy-preserving machine learning area. A popular method called Deep Leakage from Gradients (DLG) \cite{dlg} has been developed to extract training data by using the shared model gradients. An improved version of DLG, iDLG \cite{idlg}, resulted in a more reliable approach to extracting accurate data and perfectly reconstructing the labels. However, these methods lack generalization on model architecture and weight distribution initialization \cite{wang2020sapag}. In \cite{geiping2020inverting}, an analytical approach has been developed to derive the inputs before a fully connected (FC) layer. \cite{fan2020rethinking} claimed that a convolutional layer can always be converted to an FC layer, but the gradients of the original convolutional layer are still different from the gradients of the converted FC layer, which impedes the data reconstruction. In \cite{yin2021see}, the authors developed GradInversion to recover images from noise based on given gradients. All related work assumes access to the gradient of all the \emph{weights}; in this paper, we consider attacks given access only to gradients of the \emph{activations} at the cut layer.

\paragraph{Differential Privacy (DP)} One of the methods to mitigate the effectiveness of these attacks is DP \cite{dwork2006differential}. In the paper, we experiment with both DP-SGD~\cite{abadi2016deep} and Label DP~\cite{labeldp,labeldp2}. Since SL requires the device and the server to exchange activations and gradients, it can potentially leak private label information~\cite{fel, pasquini2021unleashing}. Differential privacy\cite{dpsgd} constitutes a strong standard for privacy guarantees for algorithms on aggregate databases. 

\begin{defn}
A randomized mechanism $\mathcal{M}: \mathcal{D} \rightarrow \mathcal{R}$ with domain $\mathcal{D}$ and range $\mathcal{R}$ satisfies $(\varepsilon, \delta)$-differential privacy if for any two adjacent inputs $d,d' \in \mathcal{D}$ and for any subset of outputs $S \subseteq \mathcal{R}$ it holds that:
\begin{align*}
   Pr[\mathcal{M}(d) \in S] \le e^{\varepsilon} Pr[\mathcal{M}(d') \in S] + \delta. 
\end{align*}
\end{defn}

One of the most widely adopted methods to ensure DP is through DP-SGD \cite{dpsgd}, via norm clipping and adding noise ($\sigma$) to gradients. In our case, since only the gradients of the activations at the cut layer are shared directly from the client to the server, we only consider clipping and adding noise to this part of the whole gradient. In addition, the $\varepsilon$ can be estimated through DP-accountant in various packages. There is a trade-off between model performance and model security, and carefully tuning the noise of DP is required to ensure that the trained model is effectively protected while maintaining a reasonable model performance ~\cite{kairouz2021advances}. 

On the other hand, Label DP is also based on a randomized response algorithm to improve the robustness of the training model. It is implemented when only labels need to be protected. It operates by randomly flipping the label based on the flipping probability ($p$) during training. The corresponding privacy budget $\varepsilon$ can be estimated through the formula: $p = \frac{1}{e^{\varepsilon} +1}$.

\section{Privacy and Split Learning}
\label{sec:method}

In this section, we present an extensive attack method for SL. We assume that clients have client-side private features that they would not want to share with any third party. Also, we consider that the ground-truth labels can also be private (i.e., only known to the clients) to study label leakage too. We consider the attack scenario where an honest-but-curious server follows the regular SL protocol but intends to recover clients' private data and the ground-truth labels based on the gradients of the cut layer. We design a custom and extensive attack method, termed \emph{EXACT: \underline{Ex}haustive \underline{A}tta\underline{c}k for Spli\underline{t} Learning}. We use tabular datasets throughout the paper and experiments and will discuss the extension and future research later.

We consider a $C$ class classification problem defined over a server feature space $\mathcal{X}_{server}$, a client feature space $\mathcal{X}_{client}$ and a label space $\mathcal{Y} = [C]$, where $[C] = \{ 1,...,C \}$. We define $F_{server}$ to be the server-side function, such that $F_{server}: \mathcal{X}_{server} \rightarrow \mathbb{R}^d$, which outputs the server-side activations $a_c$. We also define the client-side $F_{client}: \mathcal{X}_{client} \times \mathbb{R}^d \rightarrow  \mathcal{S}$, which maps the client feature space and the server's output to the probability simplex $\mathcal{S}$, $\mathcal{S} = \{ \textbf{z} | \sum^L_{i=1} z_i = 1, z_i \geq 0, \forall i \in [C] \}$. Both $F_{server}$ and  $F_{client}$ are parameterized over the hypothesis class $w=(w_{server}, w_{client})$, which is the weight of the neural network. $\mathcal{L}(\textbf{w})$ is the loss function, and we assume the widely used cross-entropy loss.

In this way, the server's output ($a_c$) is the activation transmitted from the server to the client at the cut layer, and the weight of the cut layer is $w_c$. Also, the gradients transmitted from the client to the server are the gradient of the cut layer activations: $\partial \mathcal{L}/ \partial a_c$. On the other hand, the gradients of the cut layer weight $\partial \mathcal{L}/\partial w_c$ stay on the client side to finish the back-propagation and allow the weights (weights on the clients-side, including the cut layer) to be updated, as shown on the left side of Figure~\ref{fig:fig}.

\paragraph{Threat Model:}  We assume a strong attacker with access to the client-side model parameters during training. While we want to study cases where the attacker has fine-grained information, the attack model can be relaxed in cases where the secure aggregation is used, and only the final model parameters are known after the training is finished. In our scenario, we assume that an honest-but-curious server has knowledge of the server-side features, and the server-side and client-side models, which is a realistic assumption given the distributed setting of both SL and FSL. As a result, we assume that an adversary can compute the server-side outputs $a_c$ for any server-side feature values, and with the input of client-side feature, the adversary can then obtain the corresponding cut-layer gradients ($\partial \mathcal{L} / \partial a_c$) from client-side. As $\partial \mathcal{L} / \partial a_c$ depends on the private features, the client-side architecture, and the output of the server $a_c$, we want to use all available information to reconstruct the private features (Figure~\ref{fig:fig}).

\subsection{Attack Method} assumes that the private features on the client side are either categorical or can be binned/clustered into a finite number of categories. We then build a list $L$ that contains all the possible configurations of client-side private features and labels. For a given sample, the adversary can calculate $a_c$ and the gradient $\partial \mathcal{L}_i / \partial a_c$ for every possible private configuration $L[i]$. We then iterate all the configurations and then match the gradient $\partial \mathcal{L}_i / \partial a_c$ by choosing the configuration $L[i]$ that minimizes the distance to the true gradient $\partial \mathcal{L} / \partial a_c$ returned by the client. Here, we choose to use the L2 distance as the distance metric to compare the gradients. The details of the algorithm can be found in Algorithm \ref{algorithm:algo}.

\begin{algorithm}[!t]
    \small
    \caption{\small\emph{EXACT}: Let $f_1,...,f_N$ be the client-side private features; $C$ be the number of classes for the task; $a_c$ be the server output; $x(_1,...,x_N,l)_i$ be the private features and label of the client i; $\frac{\partial \mathcal{L}}{\partial a_c}$ be the gradient of cut layer activation that connected to the server; $F_{client}$ be the model on the client side with $p_{clt}$ be the client output; $F_{server}$ be the model on the server side.} \label{algorithm:algo}
    \begin{algorithmic}[0]
        
        \Procedure{\emph{EXACT}}{}
            \State Get $\frac{\partial \mathcal{L}}{\partial a_c}$ \Comment{Get the transmitted gradient}
            \State Get server output $a_c$ \Comment{Get the server's output}\
            \State $L = f_1 \times f_2 \times ... \times f_N \times L$ \Comment{list of all combinations}
            \For{$i = 1,...,  |L|$}
                \State $L[i] = (x_1,...,x_N,l)_i$ \Comment{iterate the ith combination}
                \State $p_{clt} = F_{client}(a_c,(x_1,...,x_N,l)_i)$ \Comment{Forward-prop on client model}
                \State $d_i = |\frac{\partial \mathcal{L}_i}{\partial a_c} -\frac{\partial \mathcal{L}}{\partial a_c} |$ \Comment{Compute the dist. between reconstructed and original gradient }
            \EndFor
            \State $p = \operatorname{argmin}_i \{ d_i: i = 1,..., |L|\}$ \Comment{Get the index}
            \State \Return $L[p]$ \Comment{Return the reconstructed features and label}
        \EndProcedure            
    \end{algorithmic}
\end{algorithm}

Noting that in case the search space is growing with the number of features or categories to be attacked (i.e., many private features with thousands of categorical values), a heuristic approach or smart search methods can be used to speed up convergence to a given configuration. However, it is also worth mentioning that for the attack method, our priory concern is the attack performance for evaluation of the privacy leakage rather than speed. As a reference, in the datasets used here, we could successfully reconstruct features of a given sample within 16.8 seconds. Also, similar to many existing attack methods, such as DLG \cite{dlg}, iDLG \cite{idlg}, and GradInversion \cite{yin2021see}, our method \emph{EXACT} reconstructs the private data via gradient matching. Unlike the previous methods, \emph{EXACT} does not rely on optimization steps, which often involve second derivatives computations or carefully tuned regularization terms. By iterating through all possible possibilities, \emph{EXACT} guarantee to reconstruct the most relevant private features without having convergence issues in the optimization steps.

\section{Evaluation}\label{sec:exp}

\vspace{3mm}

\subsection{Experimental Setup} \label{sec:expsetup}
We conducted comprehensive experiments on three different datasets, and the details of implementation can be found below. We conducted training in both SL and FSL ways. For FSL, we simulate the federated environment by randomly allocating 16 samples for each client. All experiments are repeated three times. 

\paragraph{Datasets:} Experiments are conducted over three datasets: \emph{Adult Income dataset} \cite{adultincome}, \emph{Bank Marketing dataset} \cite{bankdataset}, and \emph{Taobao ad-display/click dataset} \cite{taobao}. The Adult Income dataset is a classification dataset aiming to predict whether the income exceeds 50K a year based on census data. It contains $48,842$ and $14$ columns. The Bank Marketing dataset is related to the direct marketing campaigns of a Portuguese banking institution. It contains $45,211$ rows and $18$ columns ordered by date. We also conduct our experiment over a production scale ad-display/click dataset of Taobao \cite{taobao}. The dataset contains $26$ million interactions (click/non-click when an Ad was shown) and $847$ thousand items across an 8-day period. We use $90\%$ of the dataset as the training set and leave $10\%$ as the testing set. 

For the Adult Income and the Bank dataset, we randomly partition the features into server features and private client features. For the Taobao dataset, we keep the user-related features as private client features. It is worth noting that since the datasets are not pre-partitioned, our methods can work with any feature partitioned, which is shown separately in Section \ref{sec:featureanalysis}. We attack all the private features, and the particular private features for each dataset can be found in Table \ref{tab:bank}\ref{tab:adult} and \ref{tab:taobao}.

\paragraph{Model Architecture} We deploy the state-of-the-art model, DeepFM, as the classification model \cite{taobao,guo2017deepfm}. We use a learning rate of $0.01$ with Adagrad and binary cross entropy as the loss function. The default number of neurons for the DNN layer for the DeepFM is chosen to be (256,128). For the attack, we train the models in different scenarios in the SL and FSL fashion with the training set and then attack the private client features using the testing set that is not seen by the model before.

\paragraph{Baseline:} We also implement two baselines to compare the attack performance. The baselines serve as guidance to show the extra information the gradient is leaking compared to our prior ability to reconstruct these private features using only server-side information. The first baseline is to reconstruct the client's features using the server's features. The second baseline is to use the server's output $a_c$ to reconstruct the client's features. For both baselines, we use the K-nearest-neighbors algorithm (KNN), which is also a non-parametric method like our method. Since our method already outputs the configurations of the features that match the closest to the original gradient, there is no extra benefit of comparing with existing parametric or optimization-based methods. 

\paragraph{DP:} We implement both Label DP and DP as explained in Section \ref{sec:related} as mitigation and defense strategy for our attack. For Label DP, we implement flipping probability $p$ of both $0.1 (\epsilon=2.2)$ and $0.01 (\epsilon=4.6)$. For DP, we implemented DP-SGD using clip norm $C$ as half the gradient norm according to \cite{andrew2021differentially} and noise multiplier $0.01$. 

\subsection{Results}\label{sec:results}

First of all, Table \ref{tab:perf} shows the model performance for all three datasets in various setups and scenarios. As all three datasets are unbalanced in terms of classes, the table reports the AUC with or without DP noise. SL and FSL columns report the unmitigated training without DP. We also consider 
Label DP, DP, and a combination of Label DP and DP for the SL training. Both methods are explained in Section \ref{sec:related}. SL and FSL report almost the same performance, which is reasonable, as both trainings are following per mini-batch steps. For Label DP, the table reports two different flipping probabilities. As we expected, with higher probability, the performance dropped for all datasets. For DP, since the noise added to the gradient is small, the degradation of the performance compared to the normal split training is also minimal. It is worth noting that the model performance (AUC), as shown in Table \ref{tab:perf} for Taobao, is much lower than the other datasets, due to the fact that Taobao is a production scale and more difficult dataset to train with. 

\begin{table*}[!t]
    \centering
    \caption{Results (AUC) of model for each dataset on test-set in different scenarios. }
    \scalebox{0.8}{
        \begin{tabular}{c|cc|cc|c|c}
            \toprule
            \textbf{Data} & \textbf{SL} & \textbf{FSL} &\multicolumn{2}{c|}{ \textbf{Label DP}} & \textbf{DP} & \textbf{Label DP \& DP } \\
            & && $p=0.1$ & $p=0.01$  &$\sigma=0.01$ & $p,\sigma=0.01$  \\
            \midrule 
            Bank& 0.88±0.0024 &0.88±0.0023 & 0.87±0.0039 & 0.88±0.0025 & 0.87±0.0003 & 0.87±0.0017 \\
            Adult & 0.89±0.0033 &  0.89±0.0030 & 0.89±0.0037 & 0.89±0.0036 & 0.89±0.0034 & 0.89±0.0024\\ 
            Taobao & 0.66±0.0001 & 0.66±0.0001 & 0.62±0.0007 & 0.65±0.0002 & 0.65±0.0012& 0.65±0.0013\\
            
            \bottomrule
        \end{tabular}
    }
    \label{tab:perf}
\end{table*}

Then, we demonstrate the attack performance in detail in Table \ref{tab:adult}, \ref{tab:bank}, and \ref{tab:taobao}. Since all features are not balanced, we choose to report the F1 score for our attack performance instead of accuracy. 

\begin{table*}[!t]
    \centering
   
    \caption{Results (F1 scores) of the feature reconstruction attack on test-set, compared with the baselines on the Adult Income dataset. The number of categories for each feature is shown in the bracket next to each feature name. For Label in Label DP, the accuracy is reported in the bracket}
    \scalebox{0.8}{
        \begin{tabular}{c|cc|cc| c| c| cc}
            \toprule
            \textbf{Features}&\textbf{SL} & \textbf{FSL} & \multicolumn{2}{c|}{ \textbf{Label DP}} & \textbf{DP} & \textbf{Comb. }& \multicolumn{2}{c}{ \textbf{Baseline}} \\
             (Num)&& &$p=0.1$ & $p=0.01$  &&&features & output \\
             \midrule
            Gender(2) & 0.9977& 0.9990& 0.9996& 0.9996& 0.3652& 0.3430 & 0.7909& 0.7729\\
            Race(5) & 0.9878& 0.9888& 0.9777& 0.9711 & 0.1119& 0.0808 & 0.3344 & 0.2789 \\
            Relationship (6) &  0.9952& 0.9957& 0.9860& 0.9974& 0.0828& 0.0756 & 0.1998 & 0.2917  \\
            Marital(7) &  0.9912 & 0.9526& 0.9794& 0.9903 & 0.1424& 0.1241 & 0.1736 & 0.2570 \\
            \midrule
            Label(2)& 1&1 & 0.80(0.90) &0.98(0.99) & 0.5497 &  0.5558 & 0.3850 & 0.5234\\
            \bottomrule
        \end{tabular}
    }
    \label{tab:adult}
\end{table*}
\begin{table*}[!t]
    \centering
    \vspace{0.2cm}
    \caption{Results (F1 scores) of the  feature reconstruction attack on test-set, compared with the baselines on the Bank Marketing dataset. The number of categories for each feature is shown in the bracket next to each feature name. For Label in Label DP, the accuracy is reported in the bracket}
    \scalebox{0.8}{
        \begin{tabular}{c|cc|cc| c| c| cc}
            \toprule
            \textbf{Features}  &\textbf{SL} & \textbf{FSL} &\multicolumn{2}{c|}{ \textbf{Label DP}} & \textbf{DP} & \textbf{Comb. }& \multicolumn{2}{c}{ \textbf{Baseline}} \\
             (Num) && &$p=0.1$ & $p=0.01$  & &&features & output \\
             \midrule
            Martial(3) & 0.9578&0.9712 &0.9877  & 0.9800 & 0.2157 &0.3037& 0.3229&0.4343\\
            Job(12) &  0.9490& 0.9515  & 0.9780 & 0.9632 & 0.0182 & 0.0188& 0.0966 & 0.1697\\
            Education(4) &0.9499 & 0.9622 & 0.9782 & 0.9795& 0.1898& 0.1280&0.2499 & 0.2845\\
            Housing(2) & 0.9835 & 0.9808& 0.9941 & 0.9911 & 0.5975 & 0.5670& 0.7112& 0.7584\\
            Loan(2) &0.9332& 0.9418 & 0.9737& 0.9621 &  0.2656 & 0.2666& 0.0909& 0.1310\\
            Contact(3) & 0.9770&0.9716 & 0.9886 & 0.9841 &  0.2683 & 0.3859& 0.5406 & 0.6102\\
            \midrule
            Label(2) &1&1 & 0.6893 (0.90) & 0.9587(0.99) & 0.3929 & 0.3682& 0.3504 & 0.4275\\
            \bottomrule
        \end{tabular}
    }
    \label{tab:bank}
\end{table*}
\begin{table*}[!t]
    \centering
    \caption{Results (F1 scores) of the  feature reconstruction attack on test-set, compared with the baselines on the Taobao dataset. The number of categories for each feature is shown in the bracket next to each feature name. For Label in Label DP, the accuracy is reported in the brackets. $0$ F1 scores indicate that the attacker reconstructs all the label as $0$ in all cases.} 
    \scalebox{0.8}{
        \begin{tabular}{c|cc|cc| c| c| cc}
            \toprule
            \textbf{Features} &\textbf{SL} &\textbf{FSL} & \multicolumn{2}{c|}{ \textbf{Label DP}} & \textbf{DP} & \textbf{Comb.}& \multicolumn{2}{c}{ \textbf{Baseline}} \\
            (Num) & & &$p=0.1$ & $p=0.01$  & &&features & output \\
             \midrule
            Age(7) & 0.8284& 0.8135  & 0.8470 & 0.7852 & 0.0464 & 0.0001 & 0.2643 & 0.1660\\
            P-value(3)& 0.8880 & 0.8499  & 0.9178 & 0.8607 & 0.0256 &0.0256& 0.4022 & 0.3280\\
            Shopping(3)&0.9034 & 0.8582& 0.9321 & 0.8583 & 0.3065  & 0.3062& 0.4503 & 0.3281\\
            Occupation(2)&0.8036& 0.8562 & 0.8931 & 0.6523 & 0.1031 & 0.0975& 0.1151 & 0.0228\\
            \midrule
            Label(2)& 1& 1 & 0.4832(0.90) & 0.9076(0.99) & 0 & 0 &  0.0066 & 0.0326\\
            \bottomrule
        \end{tabular}
    }
    \label{tab:taobao}
\end{table*}

The first thing to notice is that with both SL and FSL training for all datasets, the label can be reconstructed perfectly, which shows the importance of applying techniques such as label DP. 

Also, for both FL and FSL, the attack performance on the private features for both adult income and bank marketing datasets are all above $0.95$, implying accurate reconstructed performance for the attack in the unmitigated setting. As for the production scale dataset Taobao, some performance dropped to be around $0.80$ but still shows good attack performance, which might be due to the fact that the model performance for Taobao is lower than the other two datasets. 

As for Label DP, the attack accuracy is reported next to the F1 score in all three tables, which shows that the accuracy is exactly the same as the flipping probability. The attack performance for private features for the Label DP case is quite similar to the performance for the standard split training in all three datasets, showing that the flipping label does not provide enough protection for private features.

DP, on the other hand, has much more impact on the attack performance compared to Label DP. As we can see from Table \ref{tab:adult}, \ref{tab:bank} and \ref{tab:taobao}, the F1 scores significantly decrease for all private features and the Labels, with some of the F1 scores near 0, even with very small noise. For example, the attack performance for the Label for the Taobao dataset is 0 in Table \ref{tab:taobao} because the attack reconstructed all the labels to be label $0$. It might be due to the fact that the dataset is extremely unbalanced, with only $5\%$ of the total sample labeled as positive. As for the \texttt{Age} features in Table \ref{tab:taobao}, the F1 score is extremely low since the attack accuracy is extremely low. 

In addition, the tables also demonstrate the attack performance if we combine both Label DP and DP. As we expected, the performance for the combination will be more similar to the DP as DP has a significant impact on the attack performance, and all results show similar performance for the combined setup and the DP-only setup. 

Lastly, for the baselines, the reconstructed performance for the baselines varies depending on the dataset. The baselines provide guidance to show the lower-bound information leakage from the server side, as we might see from the tables that the attack performance for the normal SL and training with Label DP outperforms the baselines for all datasets. However, there is no clear way to distinguish between the DP case and the baselines.

\subsection{Model Architecture}\label{sec:arch}
We also conduct experiments to show that our results are not dependent on the particular model size, as shown in Table \ref{tab:arch}. We vary the DNN layers on the client side, as the gradient of the activation at the cut layer only depends on the model architecture on the client side, to see if model architecture can impact the attack performance. As we can see from the table, all SL training without mitigation demonstrates consistent performance across all private features, and the attack performances for the label are perfect. Similarly, for the DP performance, it all degrades to a similar level across all different model architectures indicating the effectiveness of DP on the attack. 

\begin{table}[!t]
    \centering
    \vspace{0.2cm}
    \caption{Results (F1 scores) of the feature reconstruction attack on test-set on the Adult Income dataset with various architecture on the client-side model.}
    \scalebox{0.75}{
        \begin{tabular}{c|ccc| cc c}
            \toprule
            \textbf{Features}&\multicolumn{3}{c|}{ \textbf{SL}} &  \multicolumn{3}{c}{ \textbf{DP}} \\
             (Num) & (256,128) & (64,32) & (32,16)&(256,128) & (64,32) & (32,16)\\
             \midrule
            Gender(2) &0.9977 & 0.9979& 0.9987&0.3652& 0.3330 & 0.1631 \\
            Race(5) & 0.9878&0.9692&0.9830&0.1119&0.2010&0.0671 \\
            Relation(6) &0.9952&0.9894&0.9896&0.0828&0.0831&0.0674    \\
            Marital(7) & 0.9912& 0.9436& 0.9789&0.1424& 0.1233& 0.1054  \\
            \midrule
            Label(2)& 1 & 1 & 1& 0.5497 & 0.4855&0.4554\\
            \bottomrule
        \end{tabular}
    }
    \label{tab:arch}
\end{table}

\subsection{Different Features}\label{sec:featureanalysis}

First, we conduct experiments to show that our results are not dependent on the selection of private features. The results are shown in Table \ref{tab:features}. It demonstrates that our method can reconstruct the private features with unmitigated SL regardless of the partition of private features and the number of categories for each feature. If we only attack 1 private feature (Gender with 2 categories or Education with 16 categories), the reconstruction performance can be $100\%$ regardless of the number of categories of the features. It is worth noticing that the attack performance drop below $0.5$ if we consider the extreme case when we attack all categorical features, but it can still perfectly attack the ground truth label. Similar to the previous results, the attack performance degrades significantly if we incorporate DP. Also, if we attack all 7 features with DP, the label attack returns everything as label $0$, which outputs the $0$ F1 score, meaning that the gradient of the cut layer yields no useful information.

In addition, we also conduct experiments to investigate the attack effectiveness if we include non-relevant private features on the client side. `\texttt{cms group}' is the private feature that does not contribute to the model performance. With or without `\texttt{cms group}', testing AUC for the model are both $0.66$, but as we can see from Table \ref{tab:taobaofeature}, adding the non-relevant features has an impact on the attack performance, especially for the unmitigated SL case.
\begin{table*}[!t]
    \centering
    \caption{Results (F1 scores) of the feature reconstruction attack on test-set on the Adult Income dataset with various sets of private features. Feature 'Occu' is short for 'Occupation', 'Work' for 'Workclass', and 'Edu' for 'Education'. For the F1 score, $1$ means that it is perfectly equal to $1$ and $1.00$ means it is rounded up to $1.00$. '-' means that the feature is not considered as a private feature, so it is not attacked.}
    \scalebox{0.78}{
        \begin{tabular}{c|cc|cc|cc|cc|cc|cc|cc|cc|cc}
            \toprule
            \textbf{Features}& \textbf{SL} & \textbf{DP} &\textbf{SL} & \textbf{DP} & \textbf{SL} & \textbf{DP} &\textbf{SL} & \textbf{DP}&\textbf{SL} & \textbf{DP}&\textbf{SL} & \textbf{DP}&\textbf{SL} & \textbf{DP} &\textbf{SL} & \textbf{DP}&\textbf{SL} & \textbf{DP}\\
             \midrule
            Gender(2) & 1 &0.36& 1 &0.33&1.00&0.26&1.00&0.36&0.97 & 0.24&0.90 & 0.21& 0.81 & 0.51& -& -& -& -\\
            Race(3) & - & - & 1 &0.33&1&0.08&1.00&0.12& 0.77 & 0.10& 0.46& 0.09& 0.25 & 0.15& -& -& -& -\\
            Marital(7) & - & - &- & - &1.00&0.14&1.00&0.09& 0.74 & 0.12& 0.50 & 0.09& 0.25 & 0.07& -& -& -& -\\
            Relation(6) &- &- & - &- & - &- &0.99&0.14& 0.92 & 0.07& 0.70 & 0.09& 0.52 & 0.04& -& -& -&-\\
            Occu.(15)& - & - &- &- & - &- &- &-& 0.83 & 0.05& 0.62 & 0.04& 0.39 & 0.08& -& -& 1.00& 0.06\\
            Work(9) & - & - &- &- & - &- &- &- & - & -& 0.52 & 0.05& 0.26 & 0.10& -& -& -& -\\
            Edu.(16) & - & - &- &- & - &- &- &- & - & -& - & -& 0.32 & 0.03& 1& 0.08& 0.99& 0.04\\
             \midrule
            Label(2)& 1 & 0.62&1 &0.77&1&0.55&1&0.55& 1 & 0.39&1& 0.32 & 1& 0 & 1& 0.62& 1& 0.66\\
            \bottomrule
        \end{tabular}
    }
    \label{tab:features}
\end{table*} 

\begin{table}[!t]
    \centering
    \caption{Results (F1 scores) of the feature reconstruction attack on test-set on the Taobao dataset with private features with or without `cms group'. $0$ F1 scores indicate that the attacker reconstructs all the labels as $0$ in all cases.}
    \scalebox{0.8}{
        \begin{tabular}{c|cc|cc}
            \toprule
            \textbf{Features} &\textbf{SL} & \textbf{DP} & \textbf{SL} & \textbf{DP}\\
            (Num) & & $\sigma=0.01$ & & $\sigma=0.01$  \\
             \midrule
            CMS Group (12) & 0.0584 & 0.0001 & - & - \\
            Age(7) & 0.6884 & 0.0119 &  0.8284 & 0.0464\\
            P-value(3)& 0.5792&  0.0246& 0.8880& 0.0256\\
            Shopping(3)& 0.6635& 0.3076& 0.9034 & 0.3065\\
            Occupation(2)& 0.3890 & 0.1032& 0.8562 & 0.1031\\
            \midrule
            Label(2)& 1& 0 &1 & 0 \\
            \bottomrule
        \end{tabular}
    }
    \label{tab:taobaofeature}
\end{table}

\subsection{Majority Vote}\label{sec:majority}

Furthermore, we also add a variation to our attack method. Instead of choosing the feature that derives the closest distance between the reconstructed gradient and the original gradient, like mentioned Algorithm \ref{algorithm:algo}, we return the $k$ closest combination of reconstructed features and take the majority vote for each feature as the final reconstructed feature. Table \ref{tab:majority} shows the results on the Adult dataset, with different values of $k$. The results show that, as $k$ increases, the attack performance drop for both unmitigated SL training and training with DP, which shows that the attack performance is highly sensitive to the variation of gradient, and it has to match exactly to generate good reconstruction performance. 

\begin{table}[!t]
    \centering
    \caption{Results (F1 scores) of the feature reconstruction attack on test-set on the Adult Income dataset with various k-value for the majority vote.}
    \scalebox{0.75}{
        \begin{tabular}{c|ccc|ccc}
            \toprule
            \textbf{Features}& \multicolumn{3}{c|}{ \textbf{SL}} & \multicolumn{3}{c}{ \textbf{ DP}}\\
             (Num) & $k=1$ & $k=5$ & $k=10$&$k=1$ &$k=5$&$k=10$\\
             \midrule
            Gender(2) & 0.9977 & 0.8231 & 0.7760 & 0.3652 & 0.3269 & 0.2761\\
            Race(5) &  0.9878 & 0.2359 & 0.0543 & 0.1119 & 0.0786 & 0.0626\\
            Relationship(6) & 0.9952 & 0.5723 & 0.3777 & 0.0828 & 0.0759 & 0.0760 \\
            Marital(7)  &  0.9912 & 0.4956 & 0.5153 & 0.1424 & 0.1348 & 0.1375 \\
            \midrule
            Label(2)& 1 & 1 & 0.9995 & 0.5479 & 0.5351 & 0.5029 \\
            \bottomrule
        \end{tabular}
    }
    \label{tab:majority}
\end{table}

\section{Discussion}
\paragraph{Feature reconstruction:} Our experiments indicate that the gradients required for SL from the client to the server can be used to reconstruct private features and labels. Compared to our baseline ability to predict the private features from the public features, the gradients significantly improve the effectiveness of the attack method in all three datasets we experimented with.  

\paragraph{Effectiveness of DP noise:} We have studied adding a wide range of DP noise, and we observe that even a small amount is enough to mitigate the reconstruction attack while allowing the model to reach similar training performance. This is because we only need to add noise on the cut-layer returned gradients. Note that in the Federated Split Learning setting, global DP is also added on the client-side models too before releasing outside the secure aggregator to ensure that the client-side weights do not encode any private information. 

\paragraph{Studying different architectures:} While we mostly focused on a typical (fully connected) DNN architecture on the client side, in future work, we want to examine further how different architectures (e.g., CNNs, RNNs, etc) can affect the ability to reconstruct private features and the sensitivity to these DP mitigation strategies. 

\paragraph{Speeding up the attack:} In our attack, we need to compare the returned gradient with every possible gradient that can result from different configurations of private features. Obviously, the wall clock time of the search depends largely on the search space, and it will significantly increase if the number of categories to search over is big. For example, for our dataset, on average, we require 20 milliseconds to finish one forward and backward pass. In this case, for the Adult dataset, each reconstruction goes over 840 times, which amounts to 16.8 seconds for each sample reconstruction. There are multiple heuristic-based grid-search techniques that can help us accelerate this search, including subset searching, gradient descent, Bayesian search~\cite{chang2021bayesian}, and so on.

\paragraph{Split learning and Federated Learning:}  Our initial experiments indicate that these findings also hold in Federated Split Learning (FSL), where there is more than one client with private features participating in training. However, in our scenarios, we randomly allocate samples to each client, which represents the Independent and identically data distribution (IID) scenario. We plan to study how non-IID can cause extra difficulties either in attacking FSL~\cite{fedprox, scaffold, zerofl}. 

\paragraph{Extending to other attacks:} In this work, we focused on reconstruction attacks on categorical features. However, \emph{EXACT} can easily be extended to membership inference attacks if we binned/clustered private features into a finite number of categories. In this case, our method would be able to infer if the particular feature is from a particular cluster.

\section{Conclusion}\label{sec:conc}

In this paper, we study the potential leakage of private features and labels during split model training. We introduce a custom and extensive feature reconstruction method and apply it to various datasets. Our results indicate that the exchanged gradients do encode private information, allowing the adversary to perfectly reconstruct labels and reconstruct private features with excellent performance. We then examine how mitigation strategies such as DP-SGD and Label DP can be used to successfully mitigate these risks without affecting the training quality. As in this work, we focused on tabular data, and we would like to expand on other tasks, such as image, text, audio processing, and other model architectures in future works.

\bibliography{aaai22}

\end{document}